\documentclass[sigconf]{acmart}
\usepackage{helvet}  
\usepackage{courier}  
\usepackage{url}  
\usepackage{graphicx}  
\usepackage{array}
\usepackage{booktabs} 
\usepackage{multirow}
\usepackage{soul}

\usepackage{amssymb,amsmath}
\usepackage{epsfig,epstopdf}
\usepackage{subfigure}
\usepackage{multirow}
\usepackage{float}
\usepackage[ruled]{algorithm}
\usepackage{algorithmic}
\usepackage{hyperref}
\usepackage{comment}
\usepackage{placeins}
\usepackage{microtype}
\usepackage{balance}
\usepackage{color}
\usepackage{soul}
\usepackage[normalem]{ulem}
\usepackage{bm}
\usepackage{hyperref}
\usepackage{algorithmic}

\newcommand{\PreserveBackslash}[1]{\let\temp=\\#1\let\\=\temp}

\newcommand{\Rmnum}[1]{\expandafter\@slowromancap\romannumeral #1@}
\newcommand{\tabincell}[2]{
\begin{tabular}{@{}#1@{}}#2\end{tabular}
}
\newcolumntype{C}[1]{>{\PreserveBackslash\centering}p{#1}}
\newcolumntype{R}[1]{>{\PreserveBackslash\raggedleft}p{#1}}
\newcolumntype{L}[1]{>{\PreserveBackslash\raggedright}p{#1}}
\graphicspath{{./graphics/}}

\begin{document}


\title{Flexible Cross-Modal Hashing}

\author{Xuanwu Liu, Jun Wang}
\affiliation{
  \institution{Southwest University, China}
}
\email{alxw1007@email.swu.edu.cn}


\author{Guoxian Yu}
\affiliation{
  \institution{Southwest University, China and KAUST, SA}
}
\email{gxyu@swu.edu.cn}

\author{Carlotta Domeniconi}
\affiliation{%
  \institution{George Mason University, USA}
}
\email{carlotta@cs.gmu.edu}

\author{Xiangliang Zhang}
\affiliation{%
  \institution{KAUST, SA}
}
\email{xiangliang.zhang@kaust.edu.sa}


%
%

\renewcommand{\shortauthors}{Xuanwu Liu et al.}

\begin{abstract}
Hashing has been widely adopted for large-scale data retrieval in many domains, due to its low storage cost and high retrieval speed. Existing cross-modal hashing methods optimistically assume that the \emph{correspondence} between training samples across modalities are readily available. This assumption  is unrealistic in practical applications. In addition, these methods generally require the \emph{same} number of samples across different modalities, which restricts their flexibility.

We propose a flexible cross-modal hashing approach (FlexCMH) to learn effective hashing codes from weakly-paired data, whose correspondence across modalities are partially (or even totally) unknown.
FlexCMH first introduces a clustering-based matching strategy to explore the local structure of each cluster, and thus to find the potential correspondence between clusters (and samples therein) across modalities.  To reduce the impact of an incomplete correspondence, it jointly optimizes in a unified objective function the potential correspondence, the cross-modal hashing functions derived from the correspondence, and a hashing quantitative loss. An alternative optimization technique is also proposed to  coordinate the correspondence and hash functions, and to reinforce
the reciprocal effects of the two objectives.  Experiments on publicly multi-modal datasets show that FlexCMH achieves significantly better results than  state-of-the-art methods, and it indeed offers a high degree of flexibility for practical cross-modal hashing tasks.
\end{abstract}


\maketitle
\section{Introduction}
\label{sec:introduction}
Hashing  has attracted an increasing interest from both research and industry, due to its low storage cost and high retrieval speed with big data  \cite{Wang2016L2H,Wang2018L2H,Gong2013Iterative,Chen2014Spectral}. Hashing aims at compressing high-dimensional vectorial data into short binary codes by preserving the structure of them, and to facilitate efficient retrieval with a significantly reduced storage. Based on the index constructed from hashing codes, big data retrieval can be made in a constant or sub-linear  time \cite{Wang2014Hashing,Wang2016L2H,Wang2018L2H,Shen2015Supervised,Liu2017Large,Zhu2013Linear,Li2015Multiview}.

With the wide range of applications of the Internet of Things, rapid influxes of multi-modal data asks for efficient cross-modal hashing solutions. For example, given an image/video about a historic event, one may want to cross-modally retrieve some texts describing the event in detail. How to perform cross-modal hashing on these widely-witnessed multi-modal data becomes  then a topic of interest in hashing \cite{Kumar2011Learning,Zhang2014Large,Wang2016L2H,Wang2018L2H}. Based on using the labels of training samples or not, existing cross-modal hashing solution can be roughly divided into unsupervised ones and supervised ones. Unsupervised ones seek hash coding functions by taking into account underlying data structure, distributions, or topological information \cite{Song2013Inter,Bronstein2010Data}. And supervised (semi-supervised) approaches try to leverage supervised information (i.e., semantic labels) to improve the performance \cite{Zhang2014Large,Lin2017Cross,Gong2014A,Wang2010Semi,Chen2008Semi}.

Existing cross-modal hashing methods optimistically assume that the correspondence between samples of different modalities is known \cite{Kang2015Learning}. However, in real applications, some objects are only available in one modality, or their corresponding (or paired) objects in another modality are only partially (or even totally) unknown. This can happen, for example, when one wants to search images from text, and there are 100 images and 200 documents, and the correspondence between 50 images and 80 documents is only partially known. In other words, the image-text collection is \emph{weakly-paired}, and only the semantic labels  are shared across modalities.  To the best of our knowledge, how to flexibly learn hashing codes from the weakly-paired data is still an \emph{untouched and challenging} topic in cross-modal hashing.

Some attempts have been made to tackle the weakly-paired multi-view data \cite{zong2018multi,Lampert2010Weakly,Liu2017Weakly}. To name a few, Weakly-paired Maximum Correlation Analysis(WMCA) extends the maximum covariance analysis to the weakly-paired case by jointly learning the latent pairs and subspace for dimensionality reduction and transfer learning \cite{Lampert2010Weakly}. Multi-modal Projection Dictionary Learning (MMPDL) jointly learns  the projective dictionary and pairing matrix for the fusion classification \cite{Liu2017Weakly}. Zong \textit{et al.} \cite{zong2018multi} assume the cluster indicator vectors of two samples from two different views should be similar if they belong to the same cluster and dissimilar otherwise, and then tackle the multi-view clustering on unpaired data by nonnegative matrix factorization. Mandal \textit{et al.} \cite{Mandal2016Generalized} learn coupled dictionaries from the respective data views and sparse representation coefficients with respect to their own dictionaries. They then maximize the correlation between sample coefficients of the same class, and simultaneously minimize the correlation of different classes to seek the matching between samples and to fuse weakly-paired multi-view data. However, these approaches still handle the weakly-paired problem in a non-flexible setting. For example, WMCA requires the number of samples in different modalities to be the same, and MMPDL needs the same number of samples for each class among different modalities. These requirements are violated in many cases, where samples across different modalities are \emph{partially-paired} and the numbers of member samples of matched clusters (or classes) across modalities are \emph{not the same}.

In this paper, we propose a Flexible Cross-Modal Hashing (FlexCMH) solution (as illustrated in Fig. \ref{Fig1}) to handle partially-paired (and even completely unpaired) multi-modal data. Our main contributions are summarized as follows:
\begin{enumerate}
\item  We design a novel matching strategy that uses  centroids of clusters, the neighborhood structure of centroids, and an incomplete correspondence between samples to seek a matching between samples in different modalities. The matching strategy neither requires the same number of samples within the matched clusters, nor across different modalities. Therefore, FlexCMH can be  applied with flexibility in general cross-modal hashing settings.\\

\item We propose a unified objective function to simultaneously consider the cross-modal matching loss, the intra-modal representation loss, and the quantitative loss to learn adaptive hashing codes. We also introduce an alternative optimization technique to jointly optimize the correspondence and hash functions, and to reinforce the reciprocal effects of these two objectives. \\

\item Experiments on benchmark multi-modal datasets show that FlexCMH significantly outperforms related and  representative cross-modal hashing approaches \cite{Bronstein2010Data,Zhang2014Large,Lin2017Cross,Lampert2010Weakly,Liu2017Weakly} in weakly paired cases, and it holds a competitive performance in different open settings. 

\end{enumerate}

The rest of this paper is organized as follows. Section \ref{sec:method} introduces the objective function of FlexCMH, and its optimization. Section \ref{sec:exp} presents the experimental setup, results, and analysis. Section \ref{sec:concl} draws the conclusions and provides directions for future work.

\begin{figure*}
\centering
{\label{Fig1}\includegraphics[width=18cm,height=7.6cm]{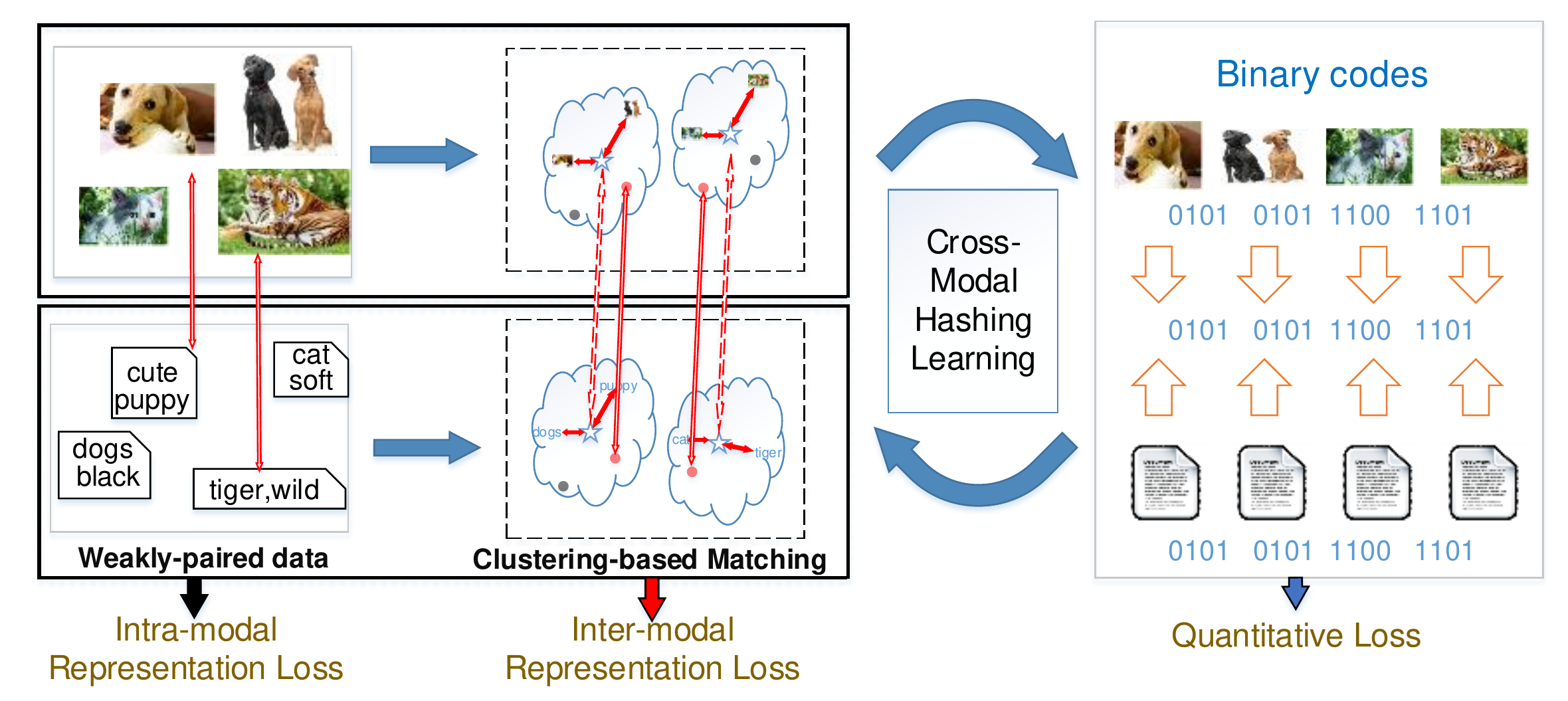}}
\vspace{-1em}
\caption{Workflow of the proposed FlexCMH (Flexible Cross-Modal Hashing). FlexCMH includes two parts: (1) A
clustering-based matching strategy to explore the matched clusters and samples therein across modalities; (2) A unified objective function to jointly account for the inter-modal representation loss,  the intra-modal representation loss, and the quantitative loss to learn adaptive hashing functions. The intra-modality presentation loss aims at exploring the   clusters and centroids of respective modalities. The inter-modal representation loss aims at preserving the proximity between samples of different modalities using matched samples. The quantitative loss aims at quantifying the hashing loss  from the high-dimensional vectors to the binary codes.}
\label{Fig1}
\end{figure*}

%
%
%
%

\section{Proposed method}
\label{sec:method}

Suppose we have $M$ modalities, and the number of training samples for the $m$-th modality is $N_m$. $\mathbf{X}^{m}\in R^{N_m\times d_m}$ represents the data matrix for the $m$-th modality, where both $N_m$ and $d_m$ are modal-dependent. $\mathbf{Y} \in \mathbb{R}^{N_m \times l}$ stores the label information of $N_m$ samples, where $l$ is the number of labels. $\mathbf{Y}_{ik}\in{\{0,1}\}$, $\mathbf{Y}_{ik}=1$ indicates that $\{\mathbf{x}_i^m\}_{m=1}^M$ is annotated with the $k$-th label; $\mathbf{Y}_{ik}=0$ otherwise. For example, in a two-modality Wiki-image search application, $\mathbf{x}^{1}_i$ is the image feature vector of sample $i$, and $\mathbf{x}^{2}_i$  is the tag vector of this sample. To enable cross-modal hashing, we need to learn two hashing functions, $F_1$: $\mathbb{R}^{d_{1}}\rightarrow {\{0,1}\}^{b}$ and $F_2$: $\mathbb{R}^{d_{2}}\rightarrow {\{0,1}\}^{b}$, where $b$ is the length of binary hash codes. These two hashing functions are expected to map $\mathbf{x}_i^{1}$ and $\mathbf{x}_i^{2}$ from the respective modality onto a common Hamming space and to preserve the proximity of the original data.

This canonical cross-modal hashing assumes that training samples in different modalities have a complete correspondence. However, the samples may be weakly-paired only. For example, consider the scenario in which, due to a temporary sensor failure, $\mathbf{x}_i^{1}$ and $\mathbf{x}_i^{2}$ do not describe the same object from different feature views. Instead, $\mathbf{x}_i^{1}$ and $\mathbf{x}_j^{2}$ ($i\neq j$) depict the same object. An intuitive solution is to only use the paired samples. However, the structure information jointly reflected by paired and unpaired samples may be distorted, thus the performance may be heavily compromised. Morevoer, if the pair information between two modalities is totally unknown, the canonical solutions cannot be applied.

To achieve an effective cross-modal hashing on such weakly-paired (or totally unpaired) multi-modal data, we introduce a flexible solution (FlexCMH), and provide its overall workflow in Fig. \ref{Fig1}. FlexCMH first introduces a clustering-based matching strategy to leverage the cluster centorids and the local structure around the centroids to explore the potential correspondence between clusters (and samples within) across different modalities. Next, it defines a permutation matrix based on the explored correspondence to unify the index of same samples across modalities. Based on the unified index, it introduces an unified objective function to simultaneously account for cross-modal similarity preserving loss,  the intra-modal representation loss and the quantitative hashing loss. An alternative optimization technique is also proposed to jointly optimize the correspondence and the hash functions, and to reinforce the reciprocal effects of these two objectives. The following subsections elaborate on the above process.

\subsection{Clustering-based cross-modal matching strategy}
Unlike single-modal hashing, the correspondence between samples is crucial for the multi-modal data fusion and retrieval. For completely matched samples,  the correspondence is completely known and can be used, along with the inter(intra)-modality similarity between samples across modalities, to learn cross-modal hashing functions. But for weakly-paired data, since the correspondence is only partially known, it's a non-trivial job to quantify the similarity between samples from different modalities.
A remedy is to divide the samples into different groups based on their labels
and impose some constraints (i.e., concerning the similarity between different classes) on the coding vectors \cite{Xiao2014Semi,Liu2018Weakly}. In the representation space, the within-class data would cluster together although they are from different modalities, and the between-class data would be placed far apart from each other. In other words, all the data vectors of the same class (different classes) from different modalities should be similar (dissimilar)\cite{wang2009multi}.
We can approximate the similarity between different classes using the centroids of respective groups\cite{Liu2017Weakly}. However, only considering centroids may not be sufficient, and the neighborhood objects around a centroid may also be helpful. Furthermore, incomplete labels of training data restrict the quality of groups.

Given  these observations, we introduce a novel clustering-based matching strategy to leverage the centroids of clusters and the local structure around the centroids. This strategy can explore the correspondence between clusters (and samples therein) between different modalities. We illustrate the clustering-based matching  strategy in the center of Fig. \ref{Fig1}, where the stars represent  centroids of  clusters in different modalities, and the red points indicate the objects with known correspondence in another modality. The likelihood that two clusters will match increases with the similarity of their centroids, with the similarity of the local structure around the centroids. To achieve that, we define a quantitative match function as follows:
\begin{equation}
\label{Eq1}
\quad  s_{c c'}^{m m'}=\sum_{g=1}^{n_s} (||\mathbf{x}_{c g}^{m}-\mathbf{z}_{c}^{m}||_F^2-\alpha||\mathbf{x}_{c'g}^{m'}-\mathbf{z}_{c'}^{m'}||_F^2)^2
\end{equation}
where $\mathbf{z}_{c}^{m}$ and $\mathbf{z}_{c'}^{m'}$ are the centriods the $c$-th cluster in the $m$-th modality and the $c'$-th cluster in the $m'$-th modality, $n_s$ is the user specified number of nearest samples of the centroids, $\mathbf{x}_{c g}^{m}$ is the $g$-th nearest sample of $\mathbf{z}_{c}^{m}$, $\alpha=||\mathbf{z}_c^{m}||_F^2/||\mathbf{z}_{c'}^{m'}||_F^2$  is a scalar coefficient to balance the scale difference between two modalities. To seek the correspondence between clusters of different modalities, Eq. (\ref{Eq1}) not only accounts for the centroids, but also for the neighborhood samples around the centroids. As such, it can explore the correspondence between neighborhood samples of respective centroids to facilitate the follow-up cross-modal hashing. In contrast, existing solutions only match centroids using labeled samples and ignore the important local patterns \cite{Lampert2010Weakly,Liu2017Weakly}. Our match function neither requires for two matched clusters to have the same number of samples, nor the same number of samples across modalities. It can also be applied to multi-modality data whose label information and correspondence are completely unknown. These advantages contribute to the flexibility of FlexCMH.

Two clusters ($c$ and $c'$) and their respective centroids $z_c^m$ and $z_{c'}^{m'}$ are matched, if $s_{c c'}^{mm'}$ is the smallest among all pairwise clusters from two modalities. We can align the objects in the respective modalities by reordering their indexes, and then use the `matched' (aligned) objects in different modalities for cross-modality hashing. To this end, we define a permutation matrix $\mathbf{\Gamma}^{mm'} \in \mathbb{R}^{N_m \times N_{m'}}$   to align samples as follows:
\begin{equation}
\label{Eq2}
\mathbf{\Gamma}_{ij}^{mm'}=\left\{
             \begin{array}{lr}
             1, & s_{c c'}^{mm'} \ \textrm{is the smallest or}\  \mathbf{P}^{mm'}_{ij}=1 \\
             0, & \textrm{otherwise}
             \end{array}
\right.
\end{equation}
where $\mathbf{P}^{mm'}_{ij}=1$ means the $i$-th sample in the $m$-th modality is  paired with the $j$-th sample in the $m'$-th modality. In this way, our cluster-based matching strategy also incorporates the known matched samples from different modalities.  $\mathbf{\Gamma}_{ij}^{mm'}=1$ if $x_i^m$ belongs to the $c$-th cluster and $x_j^{m'}$ belongs to the $c'$-th cluster, and $s_{c c'}^{mm'}$ is the smallest among all pairwise clusters from two modalities. These conditions indicate that  the indexes of $x_i^m$ and $x_j^{m'}$ should be reordered for alignment. We observe that our matching strategy is different from the typical network alignment, which aims at finding identical sub-networks \cite{Si2016FINAL,Meng2015Local}. In contrast, we aim at matching samples within the explored clusters, which describe the same object from different feature views. In addition, a sample in one modality can be paired with more than one sample in another modality. The follow-up cross-modal hashing functions can be learned using the found correspondence.

\subsection{Cross-modal hashing}
To compute the matching loss, we should first identify the centroids of the respective clusters. WMCA \cite{Liu2017Weakly} and MMPDL \cite{Liu2018Weakly} both aim at addressing cross-model learning with weakly-paired samples, but they obtain clusters using only labeled samples. In practice, the labels of samples may not be sufficient, and even unavailable. As such, they have a restricted flexibility.
To find centroids, we adopt Semi-Nonnegative Matrix Factorization (SemiNMF) \cite{Ding2010Convex} as follows:
\begin{equation}
\setlength{\abovedisplayskip}{3pt}
\label{Eq3}
\mathbf{L}_s=\sum_{m=1}^{M}||\mathbf{X}^{m}-\mathbf{Z}^{m}\mathbf{H}^{m}||^2_F, \quad s.t. \mathbf{H}^{m}\geq0
\setlength{\belowdisplayskip}{3pt}
\end{equation}
where $\mathbf{Z}^{m}\in \mathbb{R}^{d\times k}$ can be viewed as the latent representation of $k$ cluster centroids of the $m$-th modality, and $\mathbf{H}^{m} \in \mathbb{R}^{k\times N_m}$ is the soft cluster assignments of samples in the latent space. The above equation calculates the intra-modality representation loss and clustering loss simultaneously. Therefore, $\mathbf{Z}^{m}$ can be used for the clustering-based matching. $\mathbf{H}^{m}$ is the indicator matrix, which represents the probability that $N_m$ samples belong to different classes, and can be used for  hashing codes learning.

To achieve sample-to-sample cross-modal retrieval, based on the matched clusters and samples from Eq. (\ref{Eq2}), we further minimize the difference between the matched  pairs to encourage them to be as similar as possible. Specifically, the indicator vectors ($\mathbf{H}^m$) of two samples from two different modalities should be similar if they have the same cluster label, and dissimilar otherwise. To this end, we quantify the relationship between two different modalities by minimizing the deviation of the indicator vectors of pairwise objects from different modalities as follows:
\begin{equation}
\label{Eq4}
\mathbf{L}_c=\sum_{c=1}^{k}\sum_{\substack{m=1 \\ m'\neq m}}^M{||\mathbf{H}_c^m-\mathbf{H}_c^{m'}\mathbf{\Gamma}^{mm'}_{c}||_F^2}
\end{equation}
where    $\mathbf{H}_c^m \in \mathbb{R}^{N_m}$  reorders the samples in $\mathbf{X}^m$ in descending order based on their association probabilities with respect to the $c$-th class. $\mathbf{\Gamma}^{mm'}_{c} \in \mathbb{R}^{N \times N}$ is the permutation matrix, which shuffles the sample indexes in $\mathbf{H}_c^{m'}$ to align the samples according to the same indexes in  $\mathbf{H}_c^m$, which can be obtained using Eq. (\ref{Eq2}). As such, the samples of $\mathbf{H}_c^m$ can be matched with $\mathbf{H}_c^{m'}$.  In practice, we choose the top $N$  samples which belong to the $c$ ($c'$) class to setup $\mathbf{H}^m_c$ and $\mathbf{H}^{m'}_{c'}$, and to achieve cross-modal matching. As a result, our matching strategy can accommodate the case in which the number of samples belonging to the same class in different modalities can be different. In this way, we can achieve cross-modal retrieval on  multi-modal data, whose matched samples are partially or completely unknown,  even with different numbers of samples in the matched clusters.

$\mathbf{H}^{m}$ can be viewed as a soft cluster assignments of samples in the $m$-th modality with respect to $k$ clusters in a latent space. The assignments are also coordinated by the assignments in other data modalities (see Eq. (\ref{Eq4})). For cross-modal hashing, we transform the soft assignments into hard clusters $\tilde{\mathbf{H}}^{m}\in \{0|1\}^{b \times N}$ using $k$-means clustering, and then we seek the binary hash coding matrix $\mathbf{B}\in \{0|+1\}^{b \times N}$ as follows:
\begin{equation}
\label{Eq5}
\mathbf{L}_q=\sum_{m=1}^{M}||\mathbf{B}-\tilde{\mathbf{H}}^{m}||^2_F
\end{equation}
$\mathbf{B}$ can be viewed as the common Hamming space across all data modalities. It can be used for cross-modal retrieval, along with the $\mathbf{H}^{m}$ of the respective modalities. Eq. (\ref{Eq6}) is also called the hashing quantitative loss.

\subsection{Unified objective function}
Based on the above analysis, we can assemble the three losses into a unified objection function, and formulate it as:
\begin{equation}
\label{Eq6}
\begin{split}
& \mathop {\min }\limits_{\mathbf{Z}^{m}, \mathbf{H}^{m}, \mathbf{B}} \sum_{c=1}^{k}\sum_{\substack{m=1 \\ m'\neq m}}^M{||\mathbf{H}_c^m-\mathbf{H}_c^{m'}\mathbf{\Gamma}^{mm'}_{c}||_F^2}\\&+\sum_{m=1}^{M}||\mathbf{X}^{m}-\mathbf{Z}^{m}\mathbf{H}^{m}||^2_F
+\lambda\sum_{m=1}^{M}||\mathbf{B}-\tilde{\mathbf{H}}^{m}||^2_F
\end{split}
\end{equation}
where the first term quantifies the cross-modal matching loss and the inter-modal representation loss, the second term measures the intra-modal representation loss, and the third term measures the hashing code quantitative loss.  $\lambda$  is a scalar parameter that achieves a balance between the cross-modal hashing loss and the quantitative loss. By simultaneously optimizing the above three losses, we jointly account for the correspondence and the hash functions, and thus reinforce the reciprocal effects of these two objectives. This joint optimization can avoid the misleading impact of initially not well-matched clusters and samples on the subsequent cross-modal hashing. Our experimental results confirm this advantage.

\subsection{Optimization}
We observe that the loss function in Eq. (\ref{Eq6}) is actually a sum of the cross-modal matching and retrieval loss, the intra-modal representation loss, and the hashing quantitative loss. Once $\mathbf{Z}^{m}$ is fixed, we can directly obtain  $\mathbf{\Gamma}^{mm'}_{c}$ using Eq. (\ref{Eq2}).  We can solve Eq. (\ref{Eq6}) via the Alternating Direction Method of Multipliers (ADMM) \cite{Boyd2011Distributed},
which alternatively optimizes one of $\textbf{Z}^{m}$, $\textbf{H}^{m}$, and $\textbf{B}$, while keeping the other two fixed.

\textbf{Optimize $\textbf{H}^{m}$ with $\textbf{Z}^{m}$ and $\mathbf{B}$ fixed}: We utilize stochastic gradient descent (SGD) to learn $\mathbf{H}_{m}$ using the back-propagation (BP) algorithm. Here, Eq. (\ref{Eq6}) is transformed into $k$ independent optimization problems, where the $c-th$ sub-problem minimizes:
\begin{equation}
\label{Eq7}
\begin{split}
& \mathop {\min }  \sum_{\substack{m=1 \\ m'\neq m}}^M{||\mathbf{H}_c^m-\mathbf{H}_c^{m'}\mathbf{\Gamma}^{mm'}_{c}||_F^2}+\lambda\sum_{m=1}^{M}||\mathbf{X}_c^{m}-\mathbf{Z}^{m}\mathbf{H}_c^{m}||^2_F
\end{split}
\end{equation}
where $\mathbf{X}^m_c$ has the same size and samples order as $\mathbf{H}_c^{m}$. For any  class, the derivatives of Eq. (\ref{Eq7}) with respect to the indicator matrix $\mathbf{H}_c^{m}$ in the $m$ is:
\begin{equation}
\label{Eq8}
\begin{split}
\frac{\partial \mathbf{L}}{\partial \mathbf{H}_c^{m}}= 2{\mathbf{Z}^{{m}}}^{T}\mathbf{Z}^{m}\mathbf{H}_c^{m}-{\mathbf{Z}^{{m}}}^{T}\mathbf{X}_c^{m}+\lambda\sum_{ m'\neq m}^M{2(\mathbf{H}_c^{m}-\mathbf{H}_c^{m'}\mathbf{\Gamma}^{mm'}_{c})}
\end{split}
\end{equation}
We can then take $\frac{\partial \mathbf{L}}{\partial \mathbf{H}_c^{m}}$ to update the indicator matrix $\mathbf{H}_c^{m}$ using  SGD. Similarly, we can also update $\mathbf{H}_c^{m'}$ based on the derivative $\frac{\partial \mathbf{L}}{\partial \mathbf{H}_c^{m'}}$.


\textbf{Optimize $\textbf{Z}^{m}$ with $\textbf{H}^{m}$ and $\mathbf{B}$ fixed}:
Since  $\mathbf{\Gamma}^{mm'}_{c}$ depends on $\mathbf{z}_c^{m}$ and $\mathbf{z}_c^{m'}$, we compute the derivative of Eq. (\ref{Eq6}) with respect to $\mathbf{\Gamma}^{mm'}_{c}$ and $\mathbf{Z}^{m}$ as follows:
\begin{equation}
\setlength{\abovedisplayskip}{3pt}
\label{Eq10}
\begin{split}
\\&\frac{\partial \mathbf{L}}{\partial \mathbf{Z}^{m}}= \frac{\partial \mathbf{L}}{\partial \mathbf{Z}^{m}}+\frac{\partial \mathbf{L}}{\partial \mathbf{\Gamma}^{mm'}_{c}} \frac{\partial \mathbf{\Gamma}^{mm'}_{c}}{\partial \mathbf{Z}^{m}}\\&= 2\mathbf{Z}^{m}\mathbf{H}_c^{m}{\mathbf{H}_c^{m}}^T-4\lambda\mathbf{X}_c^m{\mathbf{H}_c^{m}}^T
+2{\mathbf{X}_c^{m'}\mathbf{\Gamma}^{mm'}_{c}}^T{\mathbf{H}_c^{{m'}}}^T
\end{split}
\setlength{\belowdisplayskip}{3pt}
\end{equation}
We can then use these derivatives to update the centroid matrix $\mathbf{Z}^{m}$. In each iteration, after the centroids in $\mathbf{Z}^{m}$ are updated, we consequently update $\mathbf{\Gamma}^{mm'}_{c}$ based on Eqs. (\ref{Eq1}) and (\ref{Eq2}).

\textbf{Optimize $\mathbf{B}$ with $\textbf{H}^{m}$ and  $\textbf{Z}^{m}$ fixed}: Once $\textbf{Z}^{m}$ and $\textbf{H}^{m}$ are fixed, $\tilde{\mathbf{H}}^{m}$ is also determined, then the minimization in Eq. (\ref{Eq6}) is equal to a maximization as follows:
\begin{flalign}
\label{Eq11}
\begin{split}
\mathop {\max }\limits_{\mathbf{B}} tr( \mathbf{B}^T(\lambda\sum_{m=1}^M \tilde{\mathbf{H}}^{m})\!=\!tr(\lambda\mathbf{B}^T\mathbf{U})\!=\!\sum_{i,j}\mathbf{B}_{ij}\mathbf{U}_{ij}
\end{split}
\end{flalign}
where $\mathbf{B}\in{\{-1,+1}\}^{N\times b}, \mathbf{U}=\lambda\sum_{m=1}^M \tilde{\mathbf{H}}^{m}$. It is easy to observe that the binary code $\mathbf{B}_{ij}$ should keep the same sign as $\mathbf{U}_{ij}$. Therefore, we have:
\begin{equation}
\setlength{\abovedisplayskip}{3pt}
\label{Eq12}
\mathbf{B}=sign(\mathbf{U})=sign(\lambda\sum_{m=1}^M \tilde{\mathbf{H}}^{m})
\setlength{\belowdisplayskip}{3pt}
\end{equation}
Where $sign(x)$=1 if $x>0$, $sign(x)$=0 otherwise.

By iteratively applying Eqs. (\ref{Eq8}-\ref{Eq12}), we can  jointly optimize
the correspondence and the hash functions, thus reinforcing
the reciprocal effects of these two objectives. The whole procedure of FlexCMH and the alternative optimization for solving Eq. (\ref{Eq6}) are summarized in Algorithm \ref{alg1}.

\begin{algorithm}[h!t]
\setlength{\abovedisplayskip}{3pt}
\caption{FlexCMH: Flexible Cross-Modal Hashing }
\label{alg1}
\begin{algorithmic}[1]
\REQUIRE $M$ modality data matrices $\mathbf{X}^{m}$, $m \in \{1, 2, \cdots, M\}$; the matched samples indicator matrix $\mathbf{P}^{mm'}$ (optional).
\ENSURE Clustering centroid matrices $\mathbf{Z}^{m}$ and indicator matrices $\mathbf{H}^{m}$, binary code matrix $\mathbf{B}$.
\STATE Initialize centroid matrices $\mathbf{Z}^{m}$ , indicator matrices $\mathbf{H}^{m}$, the number of classes $k$ and the number of iterations $iter$, $t=1$.
\WHILE{$t<iter$ or Eq. (\ref{Eq6}) has not converged }
 \FOR{$c=1 \to k $}
  \STATE Update $\mathbf{H}_c^{m}$ using Eqs. (\ref{Eq8})
 \ENDFOR
\STATE Update $\mathbf{Z}^{m}$ using Eq. (\ref{Eq10});
\STATE Update the permutation matrix $\mathbf{\Gamma}^{mm'}$ using Eqs. (\ref{Eq1}-\ref{Eq2}).
\STATE Update $\mathbf{B}$ using Eq. (\ref{Eq12});
\STATE $t=t+1$.
\ENDWHILE
\end{algorithmic}
\setlength{\belowdisplayskip}{3pt}
\end{algorithm}
\setlength{\belowdisplayskip}{3pt}

\subsection{Complexity analysis}
To facilitate the time complexity analysis, we assume a simple extreme case, with $M$ modalities and $k$ classes and the number of iterations is $t$. For any modality, we have $n$ samples and the extreme pairing case is considered. The time complexity of the proposed method is composed of three parts. First, the time cost of updating $H^{m}_c$ in Eq. (\ref{Eq8}) is  $O(kM(k^2d+k^2n+kdn+(k^2d)(M-1)/2))$ . Second, the  time cost of updating $Z^{m}_c $ in Eq. (\ref{Eq10}) is $O(M(4dkn+nk^2))$. Third, the  time cost of updating $\Gamma^{mm'} $ in Eq. (\ref{Eq2}) is $O(k^2n^2d^2(M(M-1))/2)$.  Since the complexity
of third part is larger than other two parts in each
iteration, the overall complexity of FlexCMH is $O(tk^2n^2d^2(M(M-1))/2)$. The empirically study (Configuration: Ubuntu 16.04.1,  Intel(R) Xeon(R)
CPU E5-2650, 256RAM) on three adopted multi-modal datasets shows that FlexCMH costs 8.532 seconds on Wiki, 43.244 seconds on Mirflickr, and 1768.196 seconds on Nus-wide.


\section{Experiments}
\label{sec:exp}
\subsection{Experimental setup}
\textbf{Datasets}:
Three widely used benchmark datasets (Nus-wide, Wiki, and Mirflicker) are collected to evaluate the performance of RDCMH. Each dataset includes two modalities, image and text, although FlexCMH can also be directly applied to cases with more than two data modalities. Nus-wide{\footnote{http://lms.comp.nus.edu.sg/research/NUS-WIDE.htm}} contains 260,648 web-text pairs. Each image is annotated with one or more labels taken from 81 concept labels. Each text is represented as a 1,000-dimensional bag-of-words vector. The hand-crafted feature of each image is a 500-dimensional bag-of-visual words (BOVW) vector. Wiki{\footnote{https://www.wikidata.org/wiki/Wikidata}} is generated from a group of 2,866 Wikipedia documents. Each document is an image-text pair, can be annotated with 10 semantic labels, and is represented by a 128-dimensional SIFT feature vector. The text articles are represented as probability distributions over 10 topics, which are derived from a Latent Dirichlet Allocation (LDA) model. Mirflickr{\footnote{http://press.liacs.nl/mirflickr/mirdownload.html}} originally contained 25,000 instances collected from Flicker. Each instance consists of an image and its associated textual tags, and is manually annotated with one or more labels, from a total of 24  semantic labels.
Each text is represented as a 1,386-dimensional bag-of-words vector, and each image is represented by a 512-dimensional GIST feature vector. 

\textbf{Comparing methods}: Six related and representative  methods are adopted for comparison. (i) CMSSH (Cross-modal Similarity Sensitive Hashing) \cite{Bronstein2010Data} treats hash code learning as a binary classification problem, and efficiently learns the hash functions using a boosting method. (ii) SCM (Semantic Correlation Maximization) \cite{Zhang2014Large} optimizes the hashing functions by maximizing the correlation between two modalities with respect to semantic labels; it includes two versions, SCM-orth and SCM-seq.  SCM-orth learns hash functions by direct eigen-decomposition with orthogonal constraints for balancing coding functions, and SCM-seq can more efficiently learn hash functions in a sequential manner without the orthogonal constraints.
(iii) CMFH (Collective matrix factorization hashing) \cite{Ding2014Collective}  learns unified binary codes using collective matrix factorization with a latent factor model on multi-modal data. (iv) SePH (Semantics Preserving Hashing) \cite{Lin2017Cross} is a probability-based hashing method, which generates one unified hash code for all
observed views by considering the semantic consistency between views. (v) WMCA (Weakly-paired Maximum Correlation Analysis) \cite{Lampert2010Weakly} adopts the maximum covariance analysis to perform the joint learning of the latent matching and subspace. (vi)MMPDL (Muti-modal projection dictionary learning) \cite{Liu2017Weakly} is a unified projective dictionary learning method, which jointly learns the projective dictionaries and matching matrix for the classification fusion. The source code of the baselines is  provided by the authors, and the input parameter values are set according to the
guidelines given by the authors in their respective papers. For WMCA and MMPDL, since they are not hashing methods, we obtain the hashing codes by exchanging the classification with the ordinary hashing function $sgn(\cdot)$.  For FlexCMH, we fix $\lambda$ in Eq. (\ref{Eq6}) to 1, $k=10$ on Wiki, $k=25$ on Mirflickr, and $k=80$ on Nus-wide;  the number of nearest neighbors $n_s$ in Eq. (\ref{Eq1}) is fixed to 5 and $N$ in Eq. (\ref{Eq4}) is fixed to 50\% of $min\{N_m, N_{m'}\}$. Our preliminary study shows FlexCMH is robust to the input values of $n_s$ and $N$. The number of iterations for optimizing Eq. (\ref{Eq6}) is set to 500. We empirically found that FlexCMH generally converges in fewer iterations on all the datasets. The parameter sensitivity of $\lambda$ and $k$ is studied in appendix.  The datasets and the code of FlexCMH will be made publicly available.



\begin{table*}
\label{Table1}
\centering
\scriptsize
\caption{
Results (MAP) on three  datasets with completely-paired data. }
\begin{tabular}{p{0.8cm}|p{1.5cm}|p{0.78cm}|p{0.78cm}|p{0.78cm}|p{0.78cm}||p{0.78cm}|p{0.78cm}|p{0.78cm}|p{0.78cm}||p{0.78cm}|p{0.78cm}|p{0.78cm}|p{0.78cm}}
\hline
 &  & \multicolumn{4}{c}{\textbf{Mirflickr}} & \multicolumn{4}{c}{\textbf{Nus-wide}} & \multicolumn{4}{c}{\textbf{Wiki}}\\\hline
 &Methods &16bits  &32bits &64bits &128bits &16bits  &32bits &64bits &128bits  &16bits  &32bits &64bits &128bits	\\\hline
 \multirow{8}{*}{\tabincell{c}{Image\\ vs.\\ Text}}
&CMSSH &	$0.5616$ &	$0.5555$ &	$0.5513$ &	$0.5484$   &	 $0.3414$ &	$0.3336$ &	$0.3282$ &	$0.3261$ & $0.1694$ &	$0.1523$ &	$0.1447$ &	$0.1434$   \\
&SCM-seq &$0.5721$ &	$0.5607$ &	$0.5535$ &	$0.5482$ &	$0.3623$ &	$0.3646$ &	$0.3703$ &	$0.3721$ & $0.1577$ &	$0.1434$ &	$0.1376$ &	$0.1358$   \\
&SCM-orth &	$0.6041$ &	$0.6112$ &	$0.6176$ &	$0.6232$  &	$0.4651$ &	$0.4714$ &	$0.4822$ &	$0.4851$ & $0.2341$ &	$0.2411$ &	$0.2443$ &	$0.2564$   \\
&CMFH &	$0.6232$ &	$0.6256$ &	$0.6268$ &	$0.6293$   & $0.4752$ &	$0.4793$ &	$0.4812$ &	$0.4866$ & $0.2578$ &	$0.2591$ &	$0.2603$ &	$0.2612$ \\
&SePH &	$0.6573$ &	$0.6603$ &	$0.6616$ &	$0.6637$   & $0.4787$ &	$0.4869$ &	$0.4888$ &	$0.4932$ & $0.2836$ &	$0.2859$ &	$0.2879$ &	$0.2863$  \\
&WMCA &$0.5834 $ &	$0.5847 $ &	$0.5856 $ &	$0.5873 $ & $0.4396 $ &	$0.4415 $ &	$0.4433 $ &	$0.4436 $  &  $0.2243 $ &	$0.2271 $ &	$0.2283 $ &	$0.2312 $ \\
&MMPDL &	$0.6126 $ &	$0.6135 $ &	$0.6141 $ &	$0.6128 $  &	 $0.4635 $ &	$0.4658 $ &	$0.4661 $ &	$0.4672 $ & $ 0.2731$ &	$0.2745 $ &	$ 0.2768$ &	$0.2801 $   \\

&FlexCMH &	${\mathbf{0.6639 }}$ &	${\mathbf{0.6674 }}$ &		${\mathbf{0.6691 }}$ & ${\mathbf{0.6724 }}$&	${\mathbf{0.4901 }}$ &		${\mathbf{0.4935 }}$ &	${\mathbf{0.4987 }}$ &${\mathbf{0.5012 }}$ & ${\mathbf{0.2846 }}$ &		${\mathbf{0.2889 }}$ &	${\mathbf{0.2912 }}$ &${\mathbf{0.2935 }}$\\
\hline

  \multirow{8}{*}{\tabincell{c}{Text \\ vs. \\ Image}}
&CMSSH &	 $0.5616$ &	$0.5551$ &	$0.5506$ &	$0.5475$  &	 $0.3392$ &	$0.3321$ &	$0.3272$ &	$0.3256$ &  $0.1578$ &	$0.1384$ &	$0.1331$ &	$0.1256$ \\
&SCM-seq &	 $0.5694$ &	$0.5611$ &	$0.5544$ &	$0.5497$  &	$0.3412$ &	$0.3459$ &	$0.3472$ &	$0.3539$ & $0.1521$ & $0.1561$ &	$0.1371$ &	$0.1261$  \\
&SCM-orth &	$0.6055$ &	$0.6154$ &	$0.6238$ &	$0.6299$    &	$0.4370$ &	$0.4428$ &	$0.4504$ &	$0.2235$  & $0.2257$ &	$0.2459$ &	$0.2482$ &	$0.2518$ \\
&CMFH &	$0.6205$ &	$0.6237$ &	$0.6259$ &	$0.6286$   & $0.4349$ &	$0.4387$ &	$0.4412$ &	$0.4425$ & $0.2872$ &	$0.2891$ &	$0.2907$ &	$0.2923$ \\
&SePH & $0.6481$ &	$0.6521$ &	$0.6545$ &	$0.6534$    & $0.4489$ &	$0.4539$ &	$0.4587$ &	$0.4621$ &  ${\mathbf{0.5345}}$ &	${\mathbf{0.5351}}$ &	${\mathbf{0.5471}}$ &	${\mathbf{0.5506}}$ \\
&WMCA & $0.5847 $ &	$0.5861 $ &	$0.5886 $ &	$0.5903 $   & $0.4179 $ &	$0.4192 $ &	$0.4221 $ &	$0.4235 $ & $0.2089 $ &	$0.2104 $ &	$0.2131 $ &	$0.2156 $  \\
&MMPDL &	 $0.6124 $ &	$0.6142 $ &	$0.6156 $ &	$0.6172 $   &	$ 0.4225$ &	$0.4232 $ &	$0.4237 $ &	$0.4256 $ &  $0.2821 $ &	$0.2824 $ &	$0.2836 $ &	$0.2861 $   \\

&FlexCMH &	${\mathbf{0.6601 }}$ &		${\mathbf{0.6632 }}$ &	${\mathbf{0.6648 }}$ &${\mathbf{0.6676 }}$&	${\mathbf{0.4639 }}$ &		${\mathbf{0.4653 }}$ &	${\mathbf{0.4688 }}$ &${\mathbf{0.4712 }} $&	${{0.2812 }}$ &		${{0.2836}}$ &	${{0.2857 }}$ &${{0.2869 }}$\\\hline
\end{tabular}
\label{Table1}
\end{table*}

\begin{table*}
\label{Table2}
\centering
\scriptsize
\caption{
Results (MAP) on three datasets with weakly-paired data. }
\begin{tabular}{p{0.8cm}|p{1.5cm}|p{0.78cm}|p{0.78cm}|p{0.78cm}|p{0.78cm}||p{0.78cm}|p{0.78cm}|p{0.78cm}|p{0.78cm}||p{0.78cm}|p{0.78cm}|p{0.78cm}|p{0.78cm}}

\hline
 &  & \multicolumn{4}{c}{\textbf{Mirflickr}} & \multicolumn{4}{c}{\textbf{Nus-wide}} & \multicolumn{4}{c}{\textbf{Wiki}}\\
 \cline{2-14}
 &Methods &16bits  &32bits &64bits &128bits &16bits  &32bits &64bits &128bits  &16bits  &32bits &64bits &128bits	\\ \cline{2-14}
 & \multicolumn{13}{c}{50\% image-text pairs are paired, CMSSH, SCM and SePH only use paired data for training}\\
\cline{2-14}
 \hline
 \multirow{8}{*}{\tabincell{c}{Image\\ vs.\\ Text}}
&CMSSH &	$0.5614$ &	$0.5551$ &	$0.5512$ &	$0.5482$   &	 $0.3411$ &	$0.3337$ &	$0.3278$ &	$0.3256$ & $0.1689$ &	$0.1520$ &	$0.1436$ &	$0.1432$   \\
&SCM-seq &$0.5720$ &	$0.5606$ &	$0.5532$ &	$0.5479$ &	$0.3620$ &	$0.3644$ &	$0.3704$ &	$0.3722$ & $0.1574$ &	$0.1436$ &	$0.1374$ &	$0.1361$   \\
&SCM-orth &	$0.6037$ &	$0.6111$ &	$0.6166$ &	$0.6235$  &	$0.4648$ &	$0.4716$ &	$0.4820$ &	$0.4853$ & $0.2344$ &	$0.2410$ &	$0.2445$ &	$0.2567$   \\
&CMFH &	$0.6225$ &	$0.6249$ &	$0.6261$ &	$0.6290$   & $0.4748$ &	$0.4789$ &	$0.4805$ &	$0.4835$ & $0.2576$ &	$0.2588$ &	$0.2596$ &	$0.2608$ \\
&SePH &	$0.6571$ &	$0.6609$ &	$0.6618$ &	$0.6636$   & $0.4785$ &	$0.4862$ &	$0.4881$ &	$0.4928$ & $0.2825$ &	$0.2853$ &	$0.2881$ &	$0.2862$  \\
&WMCA &$0.5833 $ &	$0.5848 $ &	$0.5852 $ &	$0.5836 $ & $0.4378 $ &	$0.4416 $ &	$0.4429 $ &	$0.4437 $  &  $0.2246 $ &	$0.2278 $ &	$0.2281 $ &	$0.2316 $ \\
&MMPDL &	$0.6123 $ &	$0.6131 $ &	$0.6145 $ &	$0.6121 $  &	 $0.4632 $ &	$0.4654 $ &	$0.4659 $ &	$0.4677 $ & $ 0.2729$ &	$0.2743 $ &	$ 0.2765$ &	$0.2793 $   \\
&FlexCMH &	${\mathbf{0.6635 }}$ &	${\mathbf{0.6673 }}$ &		${\mathbf{0.6689 }}$ & ${\mathbf{0.6721 }}$&	${\mathbf{0.4903 }}$ &		${\mathbf{0.4936 }}$ &	${\mathbf{0.4982 }}$ &${\mathbf{0.5010 }}$ & ${\mathbf{0.2844 }}$ &		${\mathbf{0.2885 }}$ &	${\mathbf{0.2907 }}$ &${\mathbf{0.2931 }}$\\
\hline

  \multirow{8}{*}{\tabincell{c}{Text \\ vs. \\ Image}}
&CMSSH &	 $0.5612$ &	$0.5541$ &	$0.5502$ &	$0.5474$  &	 $0.3396$ &	$0.3318$ &	$0.3269$ &	$0.3253$ &  $0.1573$ &	$0.1382$ &	$0.1326$ &	$0.1253$ \\
&SCM-seq &	 $0.5696$ &	$0.5612$ &	$0.5541$ &	$0.5486$  &	$0.3408$ &	$0.3455$ &	$0.3471$ &	$0.3536$ & $0.1517$ & $0.1556$ &	$0.1361$ &	$0.1264$  \\
&SCM-orth &	$0.6053$ &	$0.6151$ &	$0.6235$ &	$0.6294$    &	$0.4364$ &	$0.4427$ &	$0.4501$ &	$0.2233$  & $0.2256$ &	$0.2456$ &	$0.2487$ &	$0.2521$ \\
&CMFH &	$0.6201$ &	$0.6233$ &	$0.6248$ &	$0.6291$   & $0.4342$ &	$0.4382$ &	$0.4403$ &	$0.4426$ & $0.2869$ &	$0.2883$ &	$0.2902$ &	$0.2915$ \\
&SePH & $0.6479$ &	$0.6516$ &	$0.6541$ &	$0.6533$    & $0.4487$ &	$0.4536$ &	$0.4585$ &	$0.4618$ &  ${\mathbf{0.5343}}$ &	${\mathbf{0.5346}}$ &	${\mathbf{0.5468}}$ &	${\mathbf{0.5494}}$ \\
&WMCA & $0.5843 $ &	$0.5857 $ &	$0.5848 $ &	$0.5889 $   & $0.4171 $ &	$0.4186 $ &	$0.4215 $ &	$0.4232 $ & $0.2086 $ &	$0.2089 $ &	$0.2128 $ &	$0.2153 $  \\
&MMPDL &	 $0.6121 $ &	$0.6138 $ &	$0.6154 $ &	$0.6167 $   &	$ 0.4221$ &	$0.4228 $ &	$0.4231 $ &	$0.4259 $ &  $0.2813 $ &	$0.2821 $ &	$0.2832 $ &	$0.2863 $   \\
&FlexCMH &	${\mathbf{0.6578 }}$ &		${\mathbf{0.6623 }}$ &	${\mathbf{0.6641 }}$ &${\mathbf{0.6672 }}$&	${\mathbf{0.4636 }}$ &		${\mathbf{0.4657 }}$ &	${\mathbf{0.4683 }}$ &${\mathbf{0.4708 }} $&	${{0.2810 }}$ &		${{0.2831}}$ &	${{0.2855 }}$ &${{0.2864 }}$\\
\hline

 &  \multicolumn{13}{c}{50\% image-text pairs are paired, all methods use all the training data}\\
\cline{2-14}
 \multirow{8}{*}{\tabincell{c}{Image\\ vs.\\ Text}}
&CMSSH &	$0.5216$ &	$0.5238$ &	$0.5244$ &	$0.5249$   &	 $0.2715 $ &	$0.2731 $ &	$0.2757 $ &	$0.2766 $ & $0.1011$ &	$0.1023$ &	$0.1035$ &	$0.1031$   \\
&SCM-seq &$0.5398$ &	$0.5401$ &	$0.5406$ &	$0.5412$ &	$0.2953 $ &	$0.2968 $ &	$ 0.2991$ &	$0.3012 $ & $0.1107$ &	$0.1112$ &	$0.1125$ &	$0.1128$   \\
&SCM-orth &	$0.5404$ &	$0.5413$ &	$0.5430$ &	$0.5442$  &	$0.3343 $ &	$0.3358 $ &	$0.3372 $ &	$0.3395 $ & $0.1126$ &	$0.1138$ &	$0.1149$ &	$0.1168$   \\
&CMFH &	$0.5405$ &	$0.5422$ &	$0.5438$ &	$0.5447$   & $0.3409$ &	$0.3428$ &	$0.3442$ &	$0.3462$ & $0.1157$ &	$0.1165$ &	$0.1179$ &	$0.1182$ \\
&SePH &	$0.5411$ &	$0.5436$ &	$0.5467$ &	$0.5501$   & $0.3561 $ &	$0.3582 $ &	$0.3610 $ &	$0.3612 $ & $0.1235$ &	$0.1267$ &	$0.1284$ &	$0.1302$  \\
&WMCA & $0.5456 $ &	$0.5463 $ &	$0.5471 $ &	$0.5489 $ & $0.3721 $ &	$0.3746 $ &	$0.3758 $ &	$0.3761 $  &  $0.1575 $ &	$0.1593 $ &	$0.1611 $ &	$0.1635 $ \\
&MMPDL &	$0.5778 $ &	$0.5792 $ &	$0.5814 $ &	$0.5846 $  &	 $0.4117 $ &	$0.4136 $ &	$0.4137 $ &	$0.4136 $ & $0.2342 $ &	$0.2361 $ &	$0.2375 $ &	$0.2341 $   \\
&FlexCMH &	${\mathbf{0.5867 }}$ &	${\mathbf{0.5891 }}$ &		${\mathbf{0.5925 }}$ & ${\mathbf{0.5973 }}$&	${\mathbf{0.4273 }}$ &		${\mathbf{0.4296 }}$ &	${\mathbf{0.4315 }}$ &${\mathbf{0.4331 }}$ & ${\mathbf{0.2629 }}$ &		${\mathbf{0.2647 }}$ &	${\mathbf{0.2655 }}$ &${\mathbf{0.2687 }}$\\
\hline

  \multirow{8}{*}{\tabincell{c}{Text \\ vs. \\ Image}}
&CMSSH &	 $0.5121$ &	$0.5135$ &	$0.5142$ &	$0.5136$  &	 $0.2563 $ &	$0.2607 $ &	$0.2622 $ &	$0.2741 $ &  $0.0989 $ &	$0.1002$ &	$0.1011$ &	$0.1020$ \\
&SCM-seq &	 $0.5211$ &	$0.5226$ &	$0.5237$ &	$0.5242$  &	$0.2855 $ &	$0.2879 $ &	$0.2893 $ &	$0.2921 $ & $0.1118$ & $0.1124$ &	$0.1121$ &	$0.1128$  \\
&SCM-orth &$0.5235$ &	$0.5238$ &	$0.5241$ &	$0.5250$  &	$0.3211 $ &	$0.3234 $ &	$0.3269 $ &	$0.3274 $  & $0.1206$ &	$0.1209$ &	$0.1214$ &	$0.1221$ \\
&CMFH &	$0.5314 $ &	$0.5335 $ &	$0.5356 $   & $0.5372 $ &	$0.3382 $ &	$0.3397 $ &	$0.3421 $ &	$0.3442 $ & $0.1231 $ &	$0.1255 $ &	$0.1269 $ &	$ 0.1293$ \\
&SePH & $0.5431$ &	$0.5441$ &	$0.5453$ &	$0.5459$    & $0.3531 $ &	$0.3554 $ &	$0.3560 $ &	$0.3579 $ &  ${{0.1238}}$ &	${{0.1242}}$ &	${{0.1247}}$ &	${{0.1264}}$ \\
&WMCA & $0.5456 $ &	$0.5461 $ &	$0.5458 $ &	$0.5472 $   & $0.3612 $ &	$0.3648 $ &	$0.3679 $ &	$0.3712 $ & $0.1437 $ &	$0.1445 $ &	$0.1458$ &	$0.1473 $  \\
&MMPDL &	 $0.5631 $ &	$0.5647 $ &	$0.5648$ &	$0.5655 $   &	$0.3872 $ &	$0.3891 $ &	$0.3911 $ &	$0.3924 $ &  $0.2132 $ &	$0.2141 $ &	$0.2155 $ &	$0.2135 $   \\
&FlexCMH &	${\mathbf{0.5801 }}$ &		${\mathbf{0.5825 }}$ &	${\mathbf{0.5836 }}$ &${\mathbf{0.5859 }}$&	${\mathbf{0.4031 }}$ &		${\mathbf{0.4056 }}$ &	${\mathbf{0.4079 }}$ &${\mathbf{0.4112 }} $&	${\mathbf{0.2538 }}$ &		${\mathbf{0.2541 }}$ &	${\mathbf{0.2557 }}$ &${\mathbf{0.2563 }}$\\
\hline

&  \multicolumn{13}{c}{50\% image-text pairs are paired, the number of image samples and that of text samples are different}\\
\cline{2-14}
   \multirow{3}{*}{\tabincell{c}{Image \\ vs. Text}}
  &FlexCMH(nJ) &	 $0.6121 $ &	$0.6135 $ &	$0.6167 $ &	$0.6185 $   &	$0.3878 $ &	$0.3892 $ &	$0.3905 $ &	$0.3936 $ &  $0.2231 $ &	$ 0.2245 $ &	$0.2256 $ &	$0.2273 $   \\
  &FlexCMH(nC) &	 $0.5859 $ &	$0.5886 $ &	$0.5904 $ &	$0.5927 $   &$0.3618 $ &	$0.3643 $ &	$0.3666 $ &	$0.3647 $  &  $0.2015 $ &	$ 0.2057 $ &	$0.2076 $ &	$0.2088 $   \\
  &FlexCMH &	 ${\mathbf{0.6435} }$ &	${\mathbf{0.6441}} $ & ${\mathbf{0.6453}} $ &	${\mathbf{0.6468}} $   &	${\mathbf{0.4233}} $ &	${\mathbf{0.4251}} $ &	${\mathbf{0.4267}} $ &	${\mathbf{0.4283}} $ &  ${\mathbf{0.2577}} $ &	${\mathbf{0.2593}} $ &	${\mathbf{0.2612}} $ &	${\mathbf{0.2635}} $\\
  \hline

  \multirow{3}{*}{\tabincell{c}{Text  vs. \\ Image}}
  &FlexCMH(nJ) &	 $0.6224 $ &	$0.6237 $ &	$0.6244 $ &	$0.6256 $   &	$0.4115 $ &	$0.4123 $ &	$0.4143 $ &	$0.4150 $ &  $0.2435 $ &	$ 0.2456 $ &	$0.2471 $ &	$0.2493 $   \\
  &FlexCMH(nC) &	 $0.5983 $ &	$0.6004 $ &	$0.6031 $ &	$0.6042 $   &$0.3832 $ &	$0.3845 $ &	$0.3873 $ &	$0.3907 $
  &  $0.2256 $ &	$0.2274  $ &	$0.2293 $ &	$0.2308 $   \\
  &FlexCMH &	${\mathbf{0.6589 }}$ &		${\mathbf{0.6624 }}$ &	${\mathbf{0.6643 }}$ &${\mathbf{0.6658 }}$&	${\mathbf{0.4627 }}$ &		${\mathbf{0.4635 }}$ &	${\mathbf{0.4654 }}$ &${\mathbf{0.4688 }} $&	${\mathbf{0.2803 }}$ &		${\mathbf{0.2815}}$ &	${\mathbf{0.2834 }}$ &${\mathbf{0.2842 }}$\\
  \hline
 \end{tabular}
\label{Table2}
\end{table*}

\begin{table*}
\label{Table5}
\centering
\scriptsize
\caption{
Results (MAP) on three  datasets with complete-unpaired data. }
\begin{tabular}{p{1cm}|p{1.5cm}|p{0.78cm}|p{0.78cm}|p{0.78cm}|p{0.78cm}||p{0.78cm}|p{0.78cm}|p{0.78cm}|p{0.78cm}||p{0.78cm}|p{0.78cm}|p{0.78cm}|p{0.78cm}}
\hline
 &  & \multicolumn{4}{c}{\textbf{Mirflickr}} & \multicolumn{4}{c}{\textbf{Nus-wide}} & \multicolumn{4}{c}{\textbf{Wiki}}\\\hline
  &Methods &16bits  &32bits &64bits &128bits &16bits  &32bits &64bits &128bits  &16bits  &32bits &64bits &128bits	\\\hline
   \multirow{3}{*}{\tabincell{c}{Image\\ vs. \\Text}}
 & WMCA &	$0.5214 $ &	$0.5231 $ &	$0.5245 $ &	$0.5263 $    &	$0.3559 $ &	$0.3574 $ &	$0.3591 $ &	$0.3604 $ &  $0.1276 $ &	$0.1295 $ &	$0.1310$ &	$0.1336 $   \\
  &MMPDL &	 $0.5535 $ &	$0.5542 $ &	$0.5567 $ &	$0.5588 $  &	 $0.3963 $ &	$0.3984 $ &	$0.4004 $ &	$0.4015 $ & $0.2210 $ &	$0.2231 $ &	$0.2254 $ &	$0.2268 $   \\
  &FlexCMH &	${\mathbf{0.5693 }}$ &	${\mathbf{0.5704 }}$ &		${\mathbf{0.5723}}$ & ${\mathbf{0.5749 }}$&	${\mathbf{0.4115 }}$ &		${\mathbf{0.4135 }}$ &	${\mathbf{0.4159 }}$ &${\mathbf{0.4173 }}$ & ${\mathbf{0.2511 }}$ &		${\mathbf{0.2534 }}$ &	${\mathbf{0.2548 }}$ &${\mathbf{0.2563 }}$  \\\hline
  \multirow{3}{*}{\tabincell{c}{Text \\ vs. \\Image}}
 &WMCA &$0.5256 $ &	$0.5263 $ &	$0.5278 $ &	$0.5293 $   & $0.3414 $ &	$0.3438 $ &	$0.3467 $ &	$0.3481 $ & $0.1335 $ &	$0.1344 $ &	$0.1358$ &	$0.1381 $   \\
  &MMPDL &	 $0.5489 $ &	$0.5503$ &	$0.5531$ &	$0.5547 $   &	$0.3635 $ &	$0.3678 $ &	$0.3691 $ &	$0.3713 $ &  $0.2015 $ &	$0.2038 $ &	$0.2074 $ &	$0.2098 $   \\
  &FlexCMH &	${\mathbf{0.5631 }}$ &		${\mathbf{0.5652 }}$ &	${\mathbf{0.5681 }}$ &${\mathbf{0.5694 }}$&	${\mathbf{0.4031 }}$ &		${\mathbf{0.4058 }}$ &	${\mathbf{0.4083 }}$ &${\mathbf{0.4112 }} $&	${\mathbf{0.2437 }}$ &		${\mathbf{0.2459 }}$ &	${\mathbf{0.2483 }}$ &${\mathbf{0.2501 }}$  \\\hline
 \end{tabular}
\label{Table3}
\end{table*}

\subsection{Results in different practical settings}
\label{sec:results}
To study thoroughly the performance of FlexCMH and of the comparing methods, we conduct three types of experiments: (1) completely-paired, (2) weakly-paired, and (3) completely-unpaired.
In each type of experiment, all methods  are run ten times, and we report the average MAP (mean average precision) results. Since the MAP standard deviations of all methods are quite  small (less than 2\%) on all datasets, to save space the standard deviations are not reported.  The best results are boldfaced.

For the completely-paired experiments, the clustering-based matching process of FlexCMH is excluded, and each comparing method uses all the paired samples for training (70\%) and the rest for validation (30\%). Table \ref{Table1} reports the MAP results on MIRFLICKR, NUS-WIDE, and Wiki datasets.  In the Table, `Image vs. Text' denotes the setting where the query is an image and the database is text, and vice versa for `Text vs. Image'.

For the weakly-paired experiments,  we investigate three different settings: (2a) 50\% of the image-text pairs of the training set (70\% of the whole dataset) are kept, and the other pairs are randomly shuffled; in this setting the first four comparing methods (CMSSH, SCM-Seq, SCM-Orth, and SePH) can only use paired image-text instances for training and disregard the unpaired ones, whereas the last two (WMCA and MMPDL) and FlexCMH can use all the training data. (2b) As in (2a), but both the paired (50\%) and the unpaired training samples are used to train all the comparing methods. (2c) As in (2a), all the images in the training set are used for training, but 10\% of the text samples in the training set is randomly removed. As such, the number of images is different from the number of text samples across modalities and clusters. For the setting (2c), all the  comparing methods cannot be applied, so we  only report the MAP results of our FlexCMH and its variants (FlexCMH(nJ) and FlexCMH(nC)). FlexCMH(nJ)  first  seeks the matched image-text pairs, and then executes the follow-up cross-modal hashing, without jointly optimizing the matched clusters (samples) and hashing functions.  FlexCMH(nC)
uses the label information to obtain the correspondence between samples (as done by MMPDL), instead of our proposed clustering-based matching strategy. Table \ref{Table2} reports the MAP values of the compared methods in these settings.

For the completely-unpaired experiments, besides randomly partitioning the data into training (70\%) and testing (30\%) sets, we randomly shuffle the index of images and the index of text samples in the training set. As a result, the images and the text samples are almost completely unpaired. For this type of experiments, only WMCA and MMPDL can be used for comparison. Table \ref{Table3} reports the MAP values of the three  methods.

From  Table \ref{Table1}, we can see that our FlexCMH  achieves the best performance in most cases. This is because  FlexCMH  not only jointly models the cross-modal similarity preserving loss and the intra-modal similarity preserving loss, to build a more faithfully semantic projection, but also models the quantitative loss to learn  adaptive hashing codes. We observe that SePH obtains better results for `Text vs. Image' retrieval on Wiki. This is possible because of the adaptability of its probability-based strategy on small datasets. An unexpected observation is that the performance of CMSSH and SCM-Orth decreases as the length of hash codes increases. This might be caused by the imbalance between bits in the hash codes learned by singular value or eigenvalue decomposition. These experimental results show the effectiveness of FlexCMH for the canonical cross-modal hashing, where training samples from different modalities are completely paired.

From Table \ref{Table2}, we can see that the MAP results are similar to those of Table \ref{Table1}, while only 50\% of the pairs of the training set are used for training by CMSSH, SCM, and SePH. This observation suggests that this pair fraction is sufficient to train the cross-modal hashing functions. In practice, we observed a significant reduction in the MAP values when less than 10\% of the training data is paired. We also observe that the MAP results of CMSSH, SCM, and SePH sharply reduce when all the paired and unpaired samples are used for the experiment. This is because CMSSH, SCM, and SePH are misled by `incorrectly paired' (in fact, not-paired) samples. WMCA, MMPDL and FlexCMH do not manifest such a sharp  reduction in performance. That is because they adopt different techniques to augment matched samples, which boost the performance of cross-modal hashing. In addition, FlexCMH still achieves the best performance, thanks to (1) its novel clustering-based matching approach for exploring the matched clusters and samples therein, and (2) a unified objective function to optimize, in a coordinate manner, the matching between clusters and  samples, and the cross-modal hashing functions with the matched clusters and samples.

FlexCMH holds slightly reduced results when the numbers of samples (images and texts) in different modalities are not the same, and only 50\% of the image-text is paired. In `Image vs. Text' retrieval, the MAP results of FlexCMH are generally lower than those in Table \ref{Table1}. That is because 10\% of the text samples in the text modality is removed. As a result, the retrieval results may be incorrect when using the corresponding images to query the removed text.  We also observe that the results of  FlexCMH(nJ) are inferior to those of FlexCMH. This observation proves that jointly optimizing the hashing functions, and the matched clusters and samples, enables a mutual boosting of the two objectives. Furthermore, FlexCMH(nC) is also outperformed by FlexCMH, which proves that our proposed clustering-based matching strategy can more reliably find the matching between samples across modalities.

In Table \ref{Table3}, the MAP results of all  methods are inferior to those of Tables \ref{Table1} and  \ref{Table2}. Still, FlexCMH  achieves the best results, which proves the effectiveness of FlexCMH on completely-unpaired data. From these results, we can state that  the matching information of samples across modalities is crucial for cross-modal hashing. Our clustering-based matching strategy can reliably explore paired samples, and it boosts the performance of cross-modal hashing on weakly-paired (or completely unpaired) samples.

Besides, we present some examplar cross-modal retrieved images (texts) in the supplementary file to  visually support the advantages of FlexCMH.

In summary, our experimental results prove that FlexCMH can learn the cross-modal hashing more effectively than  representative comparing methods. FlexCMH is flexible in a variety of practical settings, where the paired samples across modalities are partially available or even completely unknown, and the numbers of samples in different modalities (and matched clusters) are also different. To the best of our knowledge, no existing cross-modal hashing methods can work in these scenarios.

\subsection{Parameter sensitivity analysis}

We further explore the sensitivity of the scalar parameter $\lambda$ in Eq. (\ref{Eq6}), and report the results on three datasets in Fig. \ref{Fig2}, where the code length is fixed to 16 bits. We can see that FlexCMH is slightly sensitive to $\lambda$ when $\lambda \in [10^{-3},10^{2}]$, and achieves the best performance when $\lambda=1$. Over-weighting or under-weighting the quantitative loss has a negative impact on the performance, but not significant. In summary, an effective $\lambda$ can be easily selected for FlexCMH.

In addition, we  investigate the sensitivity of the number of clusters $k$,  and report the results in Fig. \ref{Fig3} with the code length fixed to 16 bits. We can see that FlexCMH is sensitive to $k$ and can achieve the best results with $k=10$ on Wiki, $k=25$ on Mirflickr, and $k=80$ on Nus-wide. These preferred values of $k$ are close to the number of distinct labels of the corresponding datasets. Given this, we suggest to fix $k$  around the number of labels $l$.

\begin{figure}
\vspace{-0.8cm}
\centering
{\label{Fig2}\includegraphics[width=9cm,height=2.8cm]{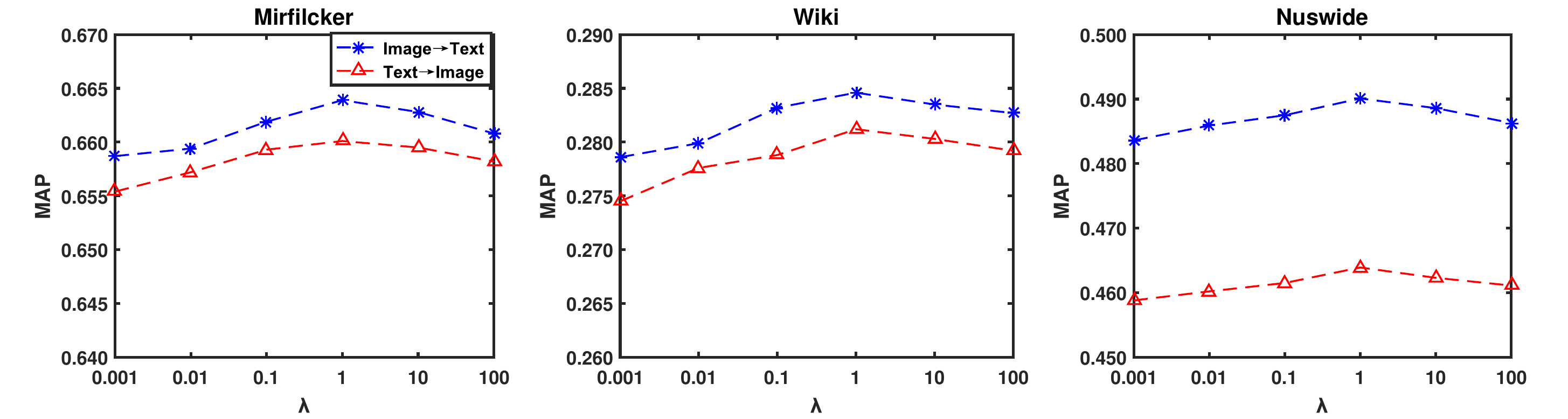}}
\caption{MAP vs. $\lambda$ on Mirfilcker and Wiki datasets.}
\label{Fig2}
\setlength{\belowcaptionskip}{-1cm}
\end{figure}

\begin{figure}
\vspace{-0.8cm}
\centering
{\label{Fig3}\includegraphics[width=9cm,height=2.8cm]{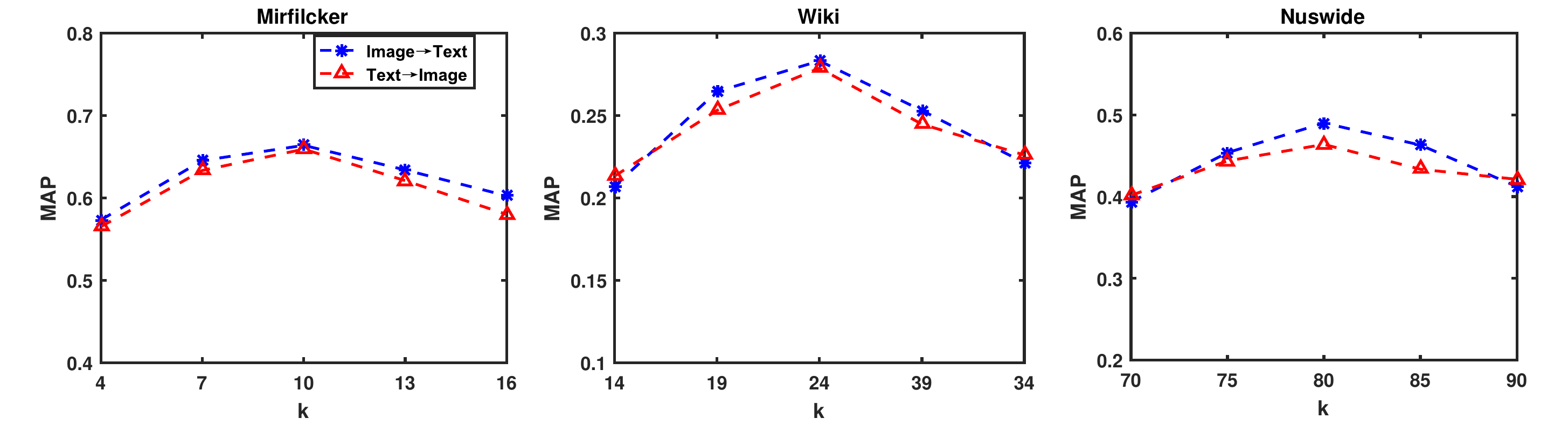}}
\caption{MAP vs. $k$ on different datasets. }
\label{Fig3}
\setlength{\belowcaptionskip}{-1cm}
\end{figure}
\subsection{Results on three modalities}
In this section, we evaluate the effectiveness of FlexCMH on the Wiki dataset  fixing code length to 16 with three  modalities.  Currently, to the best of
our knowledge, there are no  publicly available datasets with three or more modalities. To simulate a three-modality setting, we divide the 128-dimensional image modality into two sub-modalities: the 64-dim i1 and the 64-dim i2 modalities. Since the comparing methods cannot directly handle more than two modalities, we adapt them by learning hash functions between each pair of modalities, and then merge the retrieved results from the respective pairs.  For example, if Image1 serves as the query modality, then the comparing methods separately optimize two cross-modal hashing mappings (i.e., Image1 $\rightarrow$ Text and Image1 $\rightarrow$ Image2). The experimental setting is the same as in (2b). 

Table 4 shows the MAP values on the Wiki dataset
in the three-modality case. FlexCMH again outperforms the compared methods, providing evidence of the broad applicability of our proposed approach.

\begin{table}
\setlength{\abovedisplayskip}{3pt}
\label{Table7}
\centering
\caption{Results on three modalities on Wiki.}
\begin{tabular}{c|c|c|c}
\hline
  & Text  & Image1 & Image2 \\\hline
CMSSH &	0.1015 &	0.0896 & 0.0937\\
 SCM-seq & 0.1114 &	0.0975 & 0.1034\\
 SCM-orth &	0.1215	& 0.1025 & 0.1151 \\
 SePH &	0.1241	& 0.1056 & 0.1104\\
WMCA &	0.1428	& 0.1156 & 0.1241 \\
MMPDL &	0.2023	& 0.1676 & 0.1896\\
 FlexCMH &\bf 0.2331 &\bf 0.1876 &\bf 0.2031\\
 \hline
\end{tabular}\\
\label{Table4}
\setlength{\belowdisplayskip}{3pt}
\end{table}

\section{Conclusions}
\label{sec:concl}

In this paper, we  proposed a Flexible cross-modal hashing (FlexCMH) solution to learn effective hashing functions from weakly-paired (or completely-unpaired) data across modalities.  FlexCMH introduces a clustering-based matching strategy to explore the potential correspondence between clusters and their member samples.  In addition, it jointly optimizes the potential correspondence, cross-modal hashing functions derived from the correspondence and the hashing quantitative loss in a unified objective function to coordinately learn compact hashing codes. Extensive experiments demonstrate that FlexCMH outperforms the state-of-the-art hashing methods on completely-paired, weakly-paired, and completely-unpaired multi-modality data. In the future, we will incorporate deep feature learning into cross-modal hashing on weakly-paired data. The code and data (those that are not available yet) will be publicly available.

\bibliographystyle{abbrv}
\bibliography{FlexCMH_Bib}


%
%
%
%
%
%
%
%

\end{document}